\documentclass{article}

\usepackage{microtype}
\usepackage{graphicx}
\usepackage{booktabs} 
\usepackage{hyperref}

\usepackage[accepted]{icml2020_arxiv}

\icmltitlerunning{Constrained Bayesian Optimization under a Limited Budget of Failures}

\usepackage{subcaption} \usepackage{amsfonts,amsmath}
\usepackage{bm}
\usepackage{xcolor}
\usepackage{amsthm} \usepackage[capitalise]{cleveref}
\usepackage{multirow}
\usepackage{siunitx} \usepackage{tabularx}
\usepackage{xfrac} 

\newcommand{\obsf}{y}

\newcommand{\dom}{\mathcal{X}}

\newcommand{\fmin}{f_{*}}

\newcommand{\xmincons}{x^\text{c}_{*}}

\newcommand{\fminEsti}{\eta}
\newcommand{\xmin}{x_{*}}

\newcommand{\sobs}{\mathcal{D}}
\newcommand{\sobst}[1]{\mathcal{D}_t^{#1}}

\newcommand{\GP}[2]{\mathcal{GP}\left( #1, #2\right)}

\newcommand{\pnorm}[2][]{\left\lVert #2 \right\rVert_{#1}}

\newcommand{\N}[2]{\mathcal{N}( #1 ; #2 )}
\newcommand{\Nbig}[2]{\mathcal{N}\left( #1 ; #2 \right)}

\newcommand{\E}[1]{\mathbb{E}\left[ #1 \right]}
\newcommand{\Esmall}[1]{\mathbb{E}[ #1 ]}
\newcommand{\Esub}[2]{\mathbb{E}_{#1}\left[ #2 \right]}
\newcommand{\Esubsmall}[2]{\mathbb{E}_{#1}[ #2 ]}

\newcommand{\Prob}[1]{\text{Pr}(#1)}

\newcommand{\erf}[1]{\text{erf}\left( #1 \right)}

\newcommand{\prob}[1]{p( #1 )}

\newcommand{\budget}{B}
\newcommand{\evals}{T}

\newcommand{\budgetRem}{\Delta B_t}
\newcommand{\itersRem}{\Delta T_t}
\newcommand{\zsafe}{z_\text{safe}}
\newcommand{\zrisk}{z_\text{risk}}
\newcommand{\dsafe}{\rho_\text{safe}}
\newcommand{\drisk}{\rho_\text{risk}}
\newcommand{\frechet}{Fr\'{e}chet}

\newcommand{\indi}[1]{\mathbb{I} \left[ #1 \right]}

\newcommand{\PDF}[1]{\phi\left( #1 \right)}
\newcommand{\CDF}[1]{\Phi\left( #1 \right)}
\newcommand{\CDFinv}[1]{\Phi^{-1}\left( #1 \right)}

\newcommand{\R}{\mathbb{R}}

\newcommand{\xnext}{x_\text{next}}

\newcommand{\ei}{{\sc \small EI}}
\newcommand{\pim}{{\sc \small PI}}
\newcommand{\mes}{{\sc \small mES}}
\newcommand{\ucb}{{\sc \small UCB}}
\newcommand{\eic}{{\sc \small EIC}}
\newcommand{\pesc}{{\sc \small PESC}}
\newcommand{\xsearch}{{\sc \small Xs}}
\newcommand{\xsearchf}{{\sc \small XsF}}

\DeclareMathOperator*{\argmaxaux}{argmax}
\newcommand{\argmax}{\displaystyle\argmaxaux}
\DeclareMathOperator*{\argminaux}{argmin}
\newcommand{\argmin}{\displaystyle\argminaux}

\newcommand{\tableRegretBenchBothSafe}[1]{
\begin{table}[#1]
	\setlength{\tabcolsep}{4pt}
	\centering
	\caption{Constrained (top) and unconstrained benchmarks (bottom). Simple regret $r_T$ (mean $\pm$ std) and percentage of safe evaluations $\Omega$. }
	\label{tab:syn}
	\begin{tabularx}{\columnwidth}{c|cc|cc}
      {} 				& \multicolumn{2}{c}{\textsc{Hartman 6D}} 	& \multicolumn{2}{c}{\textsc{Michalewicz 10D}} \\
				\hline
			{} 				& \multicolumn{2}{c|}{$r_T$} 	& \multicolumn{2}{c}{$r_T$} \\
		\hline
		
		\ei   			& 	\multicolumn{2}{c|}{$0.75 \pm 0.00$} 			& \multicolumn{2}{c}{$0.67 \pm 0.00$} \\ 		\mes 				& 	\multicolumn{2}{c|}{$0.47 \pm 0.00$} 			& \multicolumn{2}{c}{$0.67 \pm 0.00$} \\ 		\pim 				& 	\multicolumn{2}{c|}{$0.34 \pm 0.11$} 	  	& \multicolumn{2}{c}{$0.72 \pm 0.03$} \\ 		\ucb 				& 	\multicolumn{2}{c|}{$0.39 \pm 0.18$} 	  	& \multicolumn{2}{c}{$0.70 \pm 0.06$} \\ 		\xsearch  	& 	\multicolumn{2}{c|}{$\bm{0.02 \pm 0.01}$} 	& \multicolumn{2}{c}{$\bm{0.63 \pm 0.06}$} \\ 		\hline
		\hline
			{} 				& $r_T$ 									& $\Omega\;(\%)$					& $r_T$ 								& $\Omega\;(\%)$ \\
		\hline
		\eic 		  	& 	$0.33 \pm 0.35$ 			& $68 \pm 30$				& $0.75 \pm 0.06$ 			& $15 \pm 3$ \\ 		\pesc 	  	& 	$0.14 \pm 0.22$ 			& $61 \pm 29$				& $0.74 \pm 0.07$ 			& $16 \pm 5$ \\ 		\xsearchf  	& 	$\bm{0.09 \pm 0.14}$ 	& $\bm{90 \pm 16}$ 	& $\bm{0.70 \pm 0.04}$ 	& $\bm{28 \pm 7}$ \\ 	\end{tabularx}
	\label{tab:RegretBenchBothSafe}
\end{table}
}

\newcommand{\tableHyper}[1]{
\begin{table}[#1]
	\setlength{\tabcolsep}{4pt}
	\renewcommand{\arraystretch}{1.0}
	\centering
	\caption{Hyperprior choices for the GP model hyperparameters for all experiments.}
	\begin{tabularx}{\columnwidth}{rc|cl}
      {} 													& 	& {\sc Lengthscale $\lambda$} 	& {\sc Variance $\sigma^2$} \\
		\hline
		\multirow{2}{*}{Michalewicz 10D} 	& $f$	& $\mathcal{U}(0.01,0.3)$ 					& $\mathcal{N}(0.5,0.25^2)$ \\ 
		{} 														& $g$	& $\mathcal{U}(0.01,0.3)$ 					& $\mathcal{N}(0.5,0.25^2)$ \\
		\hline
		\multirow{2}{*}{Hartman 6D} 		 	& $f$	& $\mathcal{G}(1.0,5.0)$ 					& $\mathcal{N}(0.5,0.25^2)$ \\ 
		{} 														& $g$	& $\mathcal{G}(1.0,5.0)$ 					& $\mathcal{N}(0.5,0.25^2)$ \\
				\hline
		\multirow{2}{*}{NN compression} 		& $f$	& $\mathcal{U}(0.01,0.3)$ 					& $\mathcal{N}(0.5,0.2^2)$ \\ 
		{} 														& $g$	& $\mathcal{U}(0.01,0.3)$ 					& $\mathcal{N}(7.5,2.0^2)$ \\
				\hline
		\multirow{2}{*}{Pendulum} 		& $f$	& $\mathcal{U}(0.01,0.3)$ 					& $\mathcal{N}(1.0,0.25^2)$ \\ 
		{} 														& $g$	& $\mathcal{U}(0.01,0.3)$ 					& $\mathcal{N}(0.5,0.25^2)$ \\				 
	\end{tabularx}
	\label{tab:hyper}
\end{table}
}

\newcommand{\tableRegretSynBothSafe}[1]{
\begin{table}[#1]
		\centering
	\caption{Constrained (top) and unconstrained in-model comparisons (bottom). Simple regret $r_T$ (mean $\pm$ std) and percentage of safe evaluations $\Omega$. }
	\begin{tabularx}{0.67\columnwidth}{c|cc}
      {} 				& \multicolumn{2}{c}{\textsc{3D Synthetic function}} 	\\
				\hline
			{} 				& \multicolumn{2}{c}{$r_T$} 	\\
		\hline
		
		\ei   			& 	\multicolumn{2}{c}{$1.03 \pm 0.50$} 		 \\ 		\mes 				& 	\multicolumn{2}{c}{$1.03 \pm 0.43$} 		 \\ 		\pim 				& 	\multicolumn{2}{c}{$0.86 \pm 0.41$} 	   \\ 		\ucb 				& 	\multicolumn{2}{c}{$1.00 \pm 0.43$} 	   \\ 		\xsearch  	& 	\multicolumn{2}{c}{$\bm{0.19 \pm 0.34}$}  \\ 		\hline
		\hline
			{} 				& $r_T$ 									& $\Omega\;(\%)$			\\
		\hline
		\eic 		  	& 	$0.71 \pm 0.61$ 			& $21 \pm 19$			 \\ 		\pesc 	  	& 	$1.32 \pm 0.62$ 			& $14 \pm 6$				 \\ 		\xsearchf  	& 	$\bm{0.30 \pm 0.51}$ 	& $\bm{52 \pm 15}$ 	 \\ 	\end{tabularx}
	\label{tab:RegretSynBothSafe}
\end{table}
}

 \newcommand{\figGPonedim}[1]{
\begin{figure}[#1]
\centering
\begin{subfigure}{\columnwidth}
\includegraphics[width=\columnwidth]{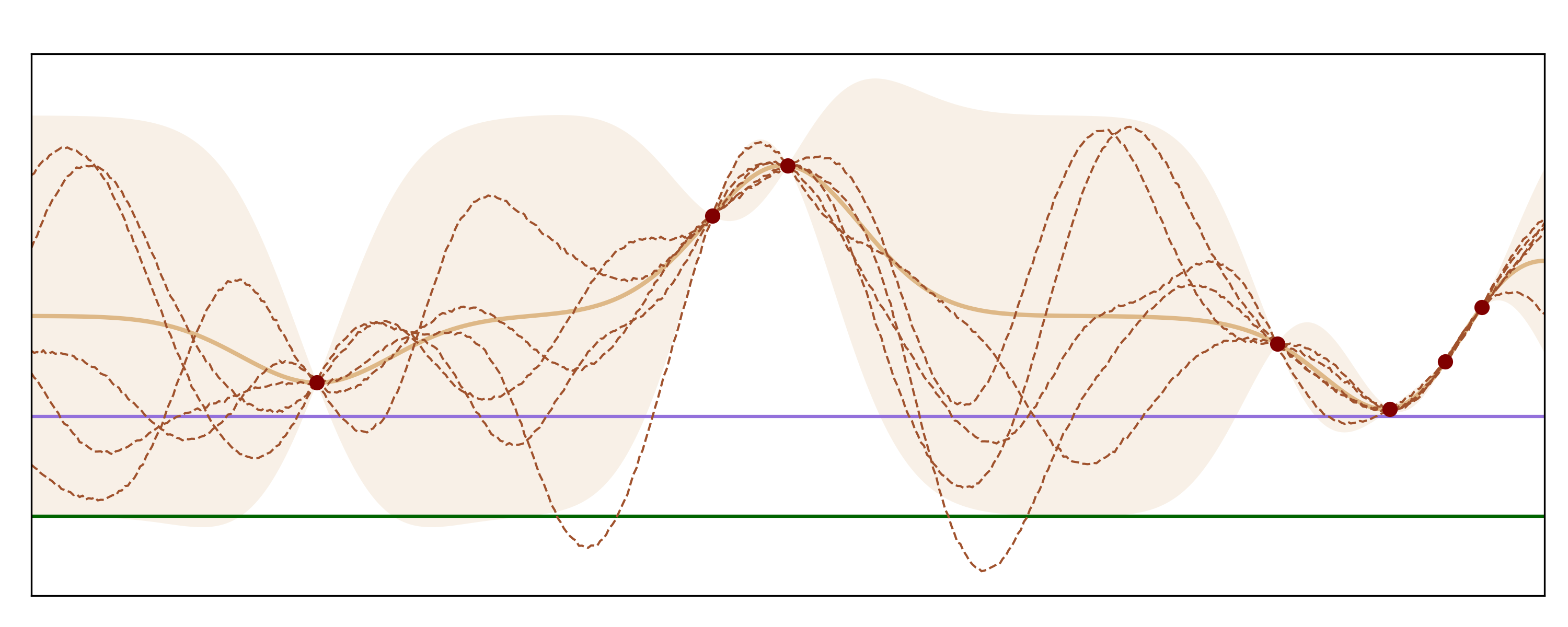}
\caption{Gaussian process posterior}
\label{sfig:excursion_GPobj}
\end{subfigure}\\
\begin{subfigure}{\columnwidth}
\includegraphics[width=\columnwidth]{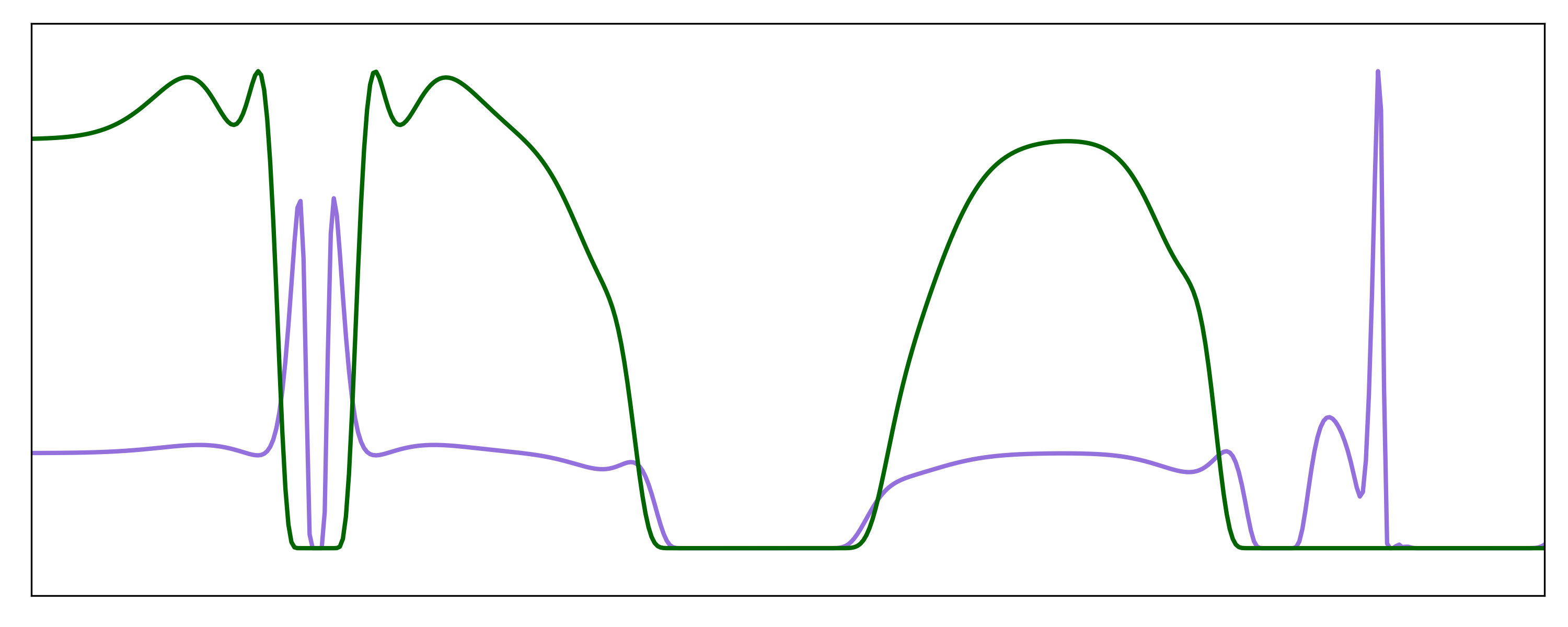}
\caption{Intensity of expected crossings $\Esmall{N_{u}(x|\sobs_t^f)}$}
\label{sfig:excursion_acqui}
\end{subfigure}\\
\caption{(a) Gaussian process posterior conditioned on a set of observations. Given a process realization (dashed lines), two choices for the threshold $u$ (solid lines) determine two different excursion sets. (b) Intensity of expected crossings $\Esmall{N_{u}(x|\sobs_t^f)}$ for each threshold $u$. Higher values correspond to areas where the boundaries of the excursion sets are likely to be, i.e., where the process is more likely to cross $u$. The curves are normalized to have the same maximum value.}
\label{fig:excursion1D}
\end{figure}
} 
\begin{document}

\twocolumn[
\icmltitle{Excursion Search for Constrained Bayesian Optimization \\ 
           under a Limited Budget of Failures}

\icmlsetsymbol{equal}{*}

\begin{icmlauthorlist}
\icmlauthor{Alonso Marco}{mpi}
\icmlauthor{Alexander von Rohr}{mpi,iav}
\icmlauthor{Dominik Baumann}{mpi}
\icmlauthor{Jos\'{e} Miguel Hern\'{a}ndez-Lobato}{cam}
\icmlauthor{Sebastian Trimpe}{mpi,aachen}
\end{icmlauthorlist}

\icmlaffiliation{mpi}{Max Planck Institute for Intelligent Systems, T\"{u}bingen, Germany}
\icmlaffiliation{cam}{Department of Engineering, University of Cambridge, Cambridge, UK}
\icmlaffiliation{aachen}{Institute for Data Science in Mechanical Engineering, RWTH Aachen University, Aachen, Germany}
\icmlaffiliation{iav}{IAV GmbH, Germany}

\icmlcorrespondingauthor{Alonso Marco}{amarco@tue.mpg.de}

\icmlkeywords{Bayesian optimization, Gaussian process, excursion sets, random fields, Rice formula, safe learning, constrained optimization}

\vskip 0.3in
]

\printAffiliationsAndNotice{}  
\allowdisplaybreaks

\begin{abstract}
When learning to ride a bike, a child falls down a number of times before achieving the first success. As falling down usually has only mild consequences, it can be seen as a tolerable failure in exchange for a faster learning process, as it provides rich information about an undesired behavior. In the context of Bayesian optimization under unknown constraints (BOC), typical strategies for safe learning explore conservatively and avoid failures by all means. On the other side of the spectrum, non conservative BOC algorithms that allow failing may fail an unbounded number of times before reaching the optimum. In this work, we propose a novel decision maker grounded in control theory that controls the amount of risk we allow in the search as a function of a given budget of failures. Empirical validation shows that our algorithm uses the failures budget more efficiently in a variety of optimization experiments, and generally achieves lower regret, than state-of-the-art methods. In addition, we propose an original algorithm for unconstrained Bayesian optimization inspired by the notion of excursion sets in stochastic processes, upon which the failures-aware algorithm is built.
\end{abstract} 
\section{Introduction}

Deploying machine learning (ML) algorithms in real-world scenarios has gained increasing interest during the last decade. Under some circumstances, lacking from sufficiently accurate models, or knowledge of the environment, such algorithms can lead to undesired outcomes. 
Deploying machine learning (ML) algorithms in real-world scenarios is an ongoing challenge. A key difficulty lies in the proper management of undesired outcomes, which are inevitable when learning under unknown or uncertain circumstances.
As an extreme case, in applications like
autonomous driving,
 a failure in the decision-making may lead to human casualties. Such safety-critical scenarios need conservative ML algorithms, which forbid any failures. 
On the other hand, there exist scenarios in which failures are still undesired, although might not come at a high cost. For example,
when deploying ML algorithms to optimize the parameters of an industrial drilling machine to drill faster, a few configurations might break the drill bits, but in exchange, a faster drilling can be learned. In such non-safety-critical applications, 
failures shall be considered as a valuable source of knowledge, and one would tolerate a limited number of them in exchange for better learning performance. 

When iteratively improving machine parameters directly from data, the mapping between a specific parameter configuration and the corresponding behavior of the machine is often unknown, and can only be revealed through experiments. Normally, such experiments are time-consuming, and thus,
  data collection is considered expensive.
In order to learn the optima of expensive black box functions, Bayesian optimization (BO) has been established in the last decade as a promising probabilistic framework \citep{shahriari2016taking}. 
Therein, the aim is to efficiently exploit the observed data in combination with prior probabilistic models to estimate the global optimum from a few trials. In the context of robot learning, BO has been used to mitigate the effort of manual controller tuning, see, e.g., \citep{calandra2016bayesian,2018vonrohr,rai2018bayesian}.

When the optimization is subject to unknown external restrictions, the goal is to solve a constrained optimization problem under multiple black box constraints.
\cite{hernandez2016general,Gelbart2014,gardner2014bayesian,gramacy2011opti,schonlau1998global,picheny2014stepwise} propose different BO methods to estimate the constrained global optimum. In \cite{NIPS2017_6785}, a variant of such problem is considered, where
the total budget of evaluations is explicitly included in the decision-making, by formulating the problem as a 
dynamic programming instance. 
Because these methods do not have a limit on the number of incurred failures, they can fail many times. In other words,
 none of them inform the decision maker about the remaining budget of failures at each iteration.

From a different perspective,
zero-budget strategies \citep{sui2015safe,berkenkamp2016safe} are needed in safety-critical applications, where failures are not allowed. Such strategies avoid failures by conservatively expanding an initially given safe area, and never exploring beyond the learned safety boundaries. However, when applied in a context where failures are allowed, such strategies become suboptimal: they will ignore such budget and miss alternative, potentially more promising, safe areas, located outisde the initial safe area.
  
In this work, we pose the problem of learning the constrained global optimum in settings where a non-zero budget of failures is given.
In particular, we make two main contributions.
Our first contribution is a failures-aware strategy for BOC
that, in contrast to prior work, does not need to be initialized in a safe region and that makes decisions taking into account the budgets of remaining failures and evaluations.

Our second contribution is a novel acquisition function inspired
by key notions of the geometry of excursion sets in stochastic processes. In \cite{adler2009random},
an excursion set is defined over smooth manifolds as those points for which a process realization crosses upwards a given threshold. The larger the threshold, the more likely it is that an \emph{upcrossing} will reveal the location of the global maximum.
Based on this intuition, we derive an acquisition function, which can be written analytically, is cheap to evaluate, and explicitly includes the process derivative to make optimal decisions.

In the following, we explain and experimentally validate the aforementioned contributions. In \cref{sec:excursion}, we characterize excursion sets in Gaussian processes (GP), and explain their benefits when used in BO. In \cref{sec:Xsearch}, we formalize the proposed novel acquisition function to solve unconstrained problems. In \cref{sec:BOfailures}, such acquisition is extended for the constrained case in the presence of a budget of failures. In \cref{sec:results}, we validate both acquisition functions empirically on common benchmarks for global optimization and real-world applications. We conclude with a discussion in \cref{sec:conclusions}.

\section{Excursion sets in Bayesian optimization}
\label{sec:excursion}
The proposed search strategy is inspired by the study of the differential and integral geometry of excursion sets in stochastic processes \citep{adler2009random}. In the particular case of GPs, analytical expressions can be derived for such sets. In the following, we provide the needed mathematical tools and intuition over which our search strategy is constructed.

\subsection{Problem formulation}
\label{sec:probuncons}
The main goal is to address the unconstrained optimization problem
\begin{equation}
\xmin = \argmin_{x\in\dom}\; f(x),
\label{eq:prob_uncons}
\end{equation}
where the objective $f : \dom \rightarrow \R$ is a black-box function, which evaluations are corrupted by noise and are expensive to collect (due to, e.g., energetic costs), and $\dom \subset \R^D$.

\subsection{Gaussian process (GP)}
\label{ssec:GPs}
We model the objective as a Gaussian process, $f \sim \GP{0}{k(x,\hat{x})}$, with covariance function $k : \dom \times \dom \rightarrow \R$, and zero prior mean. Observations $\obsf(x) = f(x) + \varepsilon$
are modeled using additive Gaussian noise $\varepsilon \sim \N{\varepsilon}{0,\sigma^2_{\text{n}}}$.
After having collected $t$ observations from the objective $\sobst{f} = \{{\bm{x}}_t , {\bm{y}}_t \} = \{ {x}_1,\dots,{x}_t,{y}_1,\dots,{y}_t \}$, its predictive distribution at a location $x$ is given by $\prob{f|\sobst{f},x} = \N{f(x)}{\mu(x|\sobst{f}),\sigma^2(x|\sobst{f})}$, with predictive mean $\mu(x|\sobst{f}) = \bm{k}^\top_t(x)[K_t + \sigma^2_{\text{n}} I ]^{-1}{\bm{y}}_t$, where the entries of vector $\bm{k}_t(x)$ are $[\bm{k}_t(x)]_i = k(x_i,x)$, the entries of the Gram matrix $K_t$ are $[K_t]_{i,j} = k({x}_i,{x}_j)$, and the entries of the vector of observations ${\bm{y}}_t$ are $[{\bm{y}}_t]_i = {y}_i$
. The predictive variance is given by $\sigma^2(x|\sobst{f}) = k(x,x) - \bm{k}_t^\top(x)[K_t + \sigma^2_{\text{n}} I ]^{-1}\bm{k}_t(x)$.
In the remainder of the paper, we drop the dependency on the current data set $\sobst{f}$ and write $\mu(x)$, $\sigma(x)$ to refer to $\mu(x|\sobst{f})$, $\sigma(x|\sobst{f})$, respectively.

\subsection{Excursion sets in Gaussian processes}
Let us assume a zero-mean scalar Gaussian process $f$, with $\dom = [0,1]^D$, $D=1$, and
 stationary covariance function $k(\tau) = k(\pnorm[2]{x - \hat{x}})$. The excursion set $\{x \in \dom: f(x)\geq u \} $ is defined as the set of locations where the process $f$ is above the threshold $u$. In \citep[Part II. Geometry]{adler2009random}, such sets are characterized by the number of upcrossings of process samples through the level $u$, i.e., $N^{+}_u = \texttt{\#} \{ x \in \dom: f(x) = u, f^\prime(x) > 0 \}$, where $f^\prime(x)$ is the derivative of the process. 
Intuitively, large $N^{+}_u$ represents a high frequency of upcrossings, which is connected with having many areas in $\dom$ where $f(x)$ lives above $u$.
 For a one-dimensional, stationary, almost surely continuous and mean-square differentiable Gaussian process,
the expected number of upcrossings \citep[Sec. 4.1]{Rasmussen2006Gaussian} is given by the well-known Rice's formula \citep[Sec. 3.1.2]{lindgren2006lectures}
\begin{align}
\E{N_u^{+}} & = \int_{0}^{1}\Esub{p(f,f^\prime|x)}{ f^\prime : f = u, f^\prime > 0} \text{d}x \label{eq:Echar_1dim} \\
& = \int_{0}^1\int_{-\infty}^{+\infty}\int_{0}^{+\infty} f^\prime\delta(f-u) p(f,f^\prime|x)\text{d}f^\prime\text{d}f\text{d}x \nonumber \\
& = \dfrac{1}{2\pi} \sqrt{ \dfrac{-k^{\prime \prime}(0)}{k(0)} }\exp\left( -\dfrac{u^2}{2k(0)} \right),\nonumber
\end{align}
where $p(f,f^\prime|x)$ is the joint density of the process and its derivative, both queried at location $x$, $\delta$ is the Dirac delta, and the second derivative of the covariance function $k^{\prime\prime}$ must exist. 
Interestingly, \eqref{eq:Echar_1dim} can be used to approximate the probability of finding the supremum of a process realization above a high level $u$. The growth rate of the approximation error with respect to $u$ is bounded 
\begin{equation}
\bigg| \E{N_u^{+}} - \Prob{ \sup_{x \in \left[0,1\right]} f(x) \geq u } \bigg| < O(e^{-\beta u^2/k(0)}),
\label{eq:sup_prob}
\end{equation}
 as $u \rightarrow \infty$, with $O(\cdot)$ indicating the limiting behavior of the approximation error and
$\beta > 1$ needs to be found \citep[Sec. 14]{adler2009random}. 
The intuitive reasoning behind this is simple:
If $f$ crosses a high level $u$, it is unlikely to do so more than once. Therefore, the probability that $f$ meets its supremum above $u$ is close to the probability that there is an upcrossing of $u$. Since the number of upcrossings of a high level will always be small, the probability of an upcrossing is well approximated by $\E{N_u^{+}}$.

While the bound in \eqref{eq:sup_prob} does not hold for the general case $D > 1$, we use it as a starting point to build 
a new acquisition function for $D \geq 1$ (cf. \cref{ssec:acquiNdim}), which shows empirically superior results than state-of-the-art BO methods.
In the following section, we show, for $D=1$, how $\E{N_u^{+}}$ can be leveraged to lead the search towards areas where the number of upcrosssings is large, or equivalently, where the global maximum is more likely to be found. Thereafter, we extend the result for $D \geq 1$.

\subsection{Practical interpretation for use in BO}
The expected number of upcrossings \eqref{eq:Echar_1dim}
 contains valuable information about the amount of times a sample realization of the process $z$ ``upcrosses'' the level $u$. However, \eqref{eq:Echar_1dim} cannot be used directly for decision-making because it is a global property of the process itself, rather than a local quantity at a specific location $x$.
 Next, we provide a practical interpretation that relaxes some of the assumptions made to obtain~\eqref{eq:Echar_1dim} and allows for its use in BO. To this end, we introduce three modifications.

First, when seeking for the optimum of the process, it is more useful to consider both, the up- and down-crossings through the level $u$, as both of them occur near the optimum when $u$ is large. This quantity is defined in \citep[Sec. 3.1.2]{lindgren2006lectures} as the expected number of \emph{crossings}
\begin{align}
\E{N_u} & = \int_{0}^{1}\Esub{p(f,f^\prime|x)}{|f^\prime| : f = u} \text{d}x \label{eq:NrCrossings} \\
& = \int_{0}^1\int_{-\infty}^{+\infty}\int_{-\infty}^{+\infty} |f^\prime|\delta(f-u) p(f,f^\prime|x)\text{d}f^\prime\text{d}f\text{d}x, \nonumber
\end{align}
with $N_u = \texttt{\#} \{ x \in \left[0,1\right]: f(x) = u\}$.
   
Second, BO uses pointwise information to decide on how interesting it is to explore a specific location $x$. 
\citep[Theorem 3.1]{lindgren2006lectures} proposes
the \emph{intensity of expected crossings} $\E{N_u(x)}$, which can be computed by simply removing the domain integral in \eqref{eq:NrCrossings}
\begin{equation}
\E{N_u(x)} = \int_{-\infty}^{+\infty}\int_{-\infty}^{+\infty} |f^\prime|\delta(f-u) p(f,f^\prime|x)\text{d}f^\prime\text{d}f.
\label{eq:IntensityCrossings}
\end{equation}

Third, when conditioning the Gaussian process $f$ on observed data $\sobs_t^f$, it becomes non-stationary\footnote{Note that all GPs are non-stationary when conditioned on data, even if the covariance function that defines them is stationary.}, and 
thus, the predictive distribution of a query $f(x)$ changes as a function of $x$. The dependency on $\sobs_t^f$ is introduced in \eqref{eq:IntensityCrossings} following \citep[Remark 3.2]{lindgren2006lectures}, as
\begin{equation}
\E{N_{u}(x|\sobs_t^f)} =  \\
\int_{-\infty}^{+\infty} |f^\prime|p(u,f^\prime|x,\sobs_t^f)\text{d}f^\prime,
\label{eq:IntensityCrossingsNonStationary}
\end{equation}
where the joint density is evaluated at $f=u$ after resolving the integral over the Dirac delta. We next provide a brief analysis for solving \eqref{eq:IntensityCrossingsNonStationary}. 

Using the rule of conditional probability, we have $p(u,f^\prime|x,\sobs_t^f) = p(u|x,\sobs_t^f) p(f^\prime|u,x,\sobs_t^f)$. The first term,
$p(u|x,\sobs_t^f) = \N{u}{\mu(x),\sigma^2(x)}$, is a Gaussian density\footnote{Using simplified notation, we write $p(u|x,\sobs_t^f)$ to refer to the density function $p_{f|x,\sobs_t^f}(\xi)$ evaluated at $\xi=u$. Similarly, we write $p(u,f^\prime|x,\sobs_t^f)$ to refer to the joint density function $p_{f,f^\prime|x,\sobs_t^f}(\xi,\zeta)$ evaluated at $\xi=u$ for some value $\zeta$.}
evaluated at $u$, with the predictive mean and variance of the GP model. The second term is also a Gaussian density over the process derivative, conditioned on $f=u$. This can be seen as adding a \emph{virtual} observation $u$ at location $x$ to existing data set. Hence, $p(f^\prime|x,u,\sobs_t^f) = p(f^\prime|\sobs_t^f \cup \{x,u\} ) = \N{f^\prime}{\mu^\prime(x),\nu^2(x)}$.
Then, \eqref{eq:IntensityCrossingsNonStationary} can be rewritten as
\begin{align}
& p(u|x,\sobs_t^f)\int_{-\infty}^{+\infty} |f^\prime|p(f^\prime|\sobs_t^f \cup \{x,u\} )\text{d}f^\prime = \label{eq:expNcrossings} \\
& \N{u}{\mu(x),\sigma^2(x)}\left( 2\nu(x) \PDF{\gamma(x)} + \mu^\prime(x) \erf{\tfrac{\gamma(x)}{\sqrt{2}}} \right), \nonumber
\end{align}
where $\gamma(x) = \mu^\prime(x) / \nu(x)$, $\phi$ is the probability density function of a standard normal distribution, and $\erf{\cdot}$ is the error function (see Appendix A for a complete derivation). \cref{fig:excursion1D} shows $\Esmall{N_{u}(x|\sobs_t^f)}$ for two different values of $u$, where the GP is conditioned on seven observations. As can be seen, different thresholds imply different intensity of crossings for the same process. When the threshold is near collected evaluations, the largest intensity of crossings tends to be concentrated near the data. On the contrary, when it is far from the data, the largest intensity of crossings is found in areas of large variance.

\figGPonedim{!t}

\subsection{Extension to $D$ dimensions}
\label{ssec:acquiNdim}
Although \eqref{eq:expNcrossings} was derived for $D=1$, we can extend it to the case $D \geq 1$. Since \eqref{eq:IntensityCrossingsNonStationary} depends on $|f^\prime|$, a natural extension is to consider the L-1 norm of the gradient of the process $\pnorm[1]{\nabla f(x)} = \sum_{j=1}^D |\tfrac{\partial f(x)}{\partial x_j}|$. Following this, we extend \eqref{eq:expNcrossings} as $\Esubsmall{p(f(x),\nabla f(x))}{\pnorm[1]{\nabla f(x)} : f(x) = u, \sobs_t^f}$, 
 \begin{flalign}
\E{N_{u}(x|\sobs_t^f)} & \simeq \N{u}{\mu(x),\sigma^2(x)} \times \label{eq:expNcrossingsNdim} \\
& \sum_{j=1}^D\left( 2\nu_j(x) \phi(\gamma_j(x)) + \mu_j(x) \erf{\tfrac{\gamma_j(x)}{\sqrt{2}}} \right), \nonumber
\end{flalign}
where $\gamma_j(x) = \mu_j(x) / \nu_j(x)$. The gradient $\nabla f(x) \sim \N{\nabla f(x)}{\nabla \mu(x),V(x)}$ follows a multivariate Gaussian, and $\mu_j(x) = \left[ \nabla \mu(x) \right]_j$ and $\nu_j(x) = (\left[ V(x) \right]_{jj})^{1/2} = \sqrt{ \partial^2k(x_j,x_j)/\partial x^2_j }$. Note that $\nabla \mu(x)$ and $V(x)$ depend on the extended data set $\sobs_t^f \cup \{x,u\}$.

In the following sections, we propose two novel algorithms that build upon the quantity \eqref{eq:expNcrossingsNdim}.
 
\section{Excursion search algorithm}
\label{sec:Xsearch}
The modifications applied to \eqref{eq:Echar_1dim}, detailed above, allow extracting useful information 
about how likely is the process $f$ to cross a certain level $u$ at each location $x$. When $u$ is a lower bound on the collected data, \eqref{eq:expNcrossingsNdim} reveals locations where the process is more likely to have a minimum. If we repeatedly evaluate at such locations, one would expect to approach faster
the global minimum. In the following, we characterize \eqref{eq:expNcrossingsNdim} as an acquisition function for optimal decision-making.

\subsection{Threshold of crossings as the global minimum}
\label{ssec:thres}
The choice of the threshold $u$ in \eqref{eq:expNcrossingsNdim} is important when trying to find the global minimum. A hypothetically appropriate low value for $u$ is right above the global minimum $f_* = f(x_*)$, i.e., $u = f_* + \epsilon$, where $\epsilon > 0$ is small. Then, if crossings through $u = f_* + \epsilon$ are likely to occur at a specific area, we know that such area is likely to contain the global minimum, and thus, will show a large $\Esmall{N_{u}(x|\sobs_t^f)}$. However, in practice we do not have access to the true $\fmin$ of the objetive function, and thus, cannot compute $u$ in the aforementioned way. At most, we are able to assume a distribution over the global minimum $\fmin \sim p(\fmin)$, implied by the GP model on $f$. In the following, we assume that $u$ follows such distribution, i.e., $u \sim p(u) = p(\fmin)$.

It is well-known in extreme value theory \citep{de2007extreme} that $\fmin$ follows one of the three extreme value distributions: Gumbel, \frechet, or Weibull, which generally model tails distributions. For example, in \cite{wang2017mes}, the Gumbel distribution is chosen to model $p(\fmin)$. However, such distribution has infinite support, while in practice it is not useful to have any probability mass above the best observed evaluation $\eta = \min( y(x_1),\ldots,y(x_T) )$. Instead, we consider the \frechet~distribution as a more appropriate choice as it provides finite support $\fmin \leq \fminEsti$. For minimization problems, we can define it in terms of its survival function $\mathcal{F}_{s,q}(a) = \Prob{\fmin \geq a}$, given by
\begin{equation}
\mathcal{F}_{s,q}(a)
=
\left\lbrace
\begin{array}{ll}
0, & \text{if } a > \fminEsti \\
\exp\left(-\left( \tfrac{\fminEsti - a}{s} \right)^{-q}\right), & \text{if } a \leq \fminEsti
\end{array}
\right.
\label{eq:FrechetCDF}
\end{equation}
where $\Prob{\fmin \geq a}=\int_{a}^{+\infty}p(\fmin)\text{d} \fmin$, and the parameters $s > 0$ and $q > 1$ can be estimated from data following the same approach as in \citep[Appendix B]{wang2017mes}. A thorough analysis on the advantage of using the Fr\'{e}chet distribution, instead of the Gumbel distribution, for gathering samples of $\fmin$ can be found in Appendix B. Using the above definition, the stochastic threshold $u \sim p(u) = p(\fmin)$, makes the quantity \eqref{eq:expNcrossingsNdim} also stochastic. We propose to compute its expectation over $u$, i.e., $\Esubsmall{p(u)}{ \Esmall{N_{u}(x|\sobs_t^f)}} = \Esubsmall{p(\fmin)}{ \Esmall{N_{\fmin}(x|\sobs_t^f)}}$, which we explain next.

\subsection{Acquisition function}
\label{ssec:acqui}
We define the \emph{excursion search} (\xsearch) acquisition function as
\begin{align}
\alpha_\text{X}(x) & = \Esub{p(\fmin)}{\E{N_{\fmin}(x|\sobs_t^f)}} \label{eq:Xsearch} \\
& \simeq \dfrac{1}{S}\sum_{l=1}^S \E{N_{\fmin^l}(x|\sobs_t^f)}, \nonumber
\end{align}

where the outer expectation is intractable and is approximated via sampling. For each sample $\fmin^l \sim p(\fmin)$, \eqref{eq:expNcrossingsNdim} needs to be recomputed.
The samples
can be collected through the inverse of \eqref{eq:FrechetCDF}, $\fmin^l = \mathcal{F}_{s,q}^{-1}(\xi^l)$. $\xi^l \sim U(0,1)$ follows a uniform distribution in the unit interval, and $\mathcal{F}_{s,q}^{-1}(\xi^l) = \eta - s(-\log(1-\xi^l))^{-1/q}$.

Intuitively, the \xsearch~acquisition function \eqref{eq:Xsearch} reveals areas near the global maximum (i.e., where the gradient crosses the estimated $\fmin$ with large norm), instead of directly aiming at potential maximums, minimums, or saddle points.
Furthermore, \xsearch~inherently trades off exploration with exploitation: At early stages of the search, the estimated Fr\'{e}chet distribution \eqref{eq:FrechetCDF} reflects large uncertainty about $\fmin$, which causes the samples $\fmin^l$ to lie far from the data. Hence, exploration is encouraged, as shown in \cref{fig:excursion1D} (green lines). At later stages, when more data is available, the Fr\'{e}chet distribution \eqref{eq:FrechetCDF} shrinks toward the lowest observations, which then encourages exploitation, as shown in \cref{fig:excursion1D} (violet lines).

The acquisition \eqref{eq:Xsearch} is our first contribution, and can be used for unconstrained optimization problems, e.g., \eqref{eq:prob_uncons}.

 \section{Bayesian optimization with a limited budget of failures}
\label{sec:BOfailures}
In the previous section, we introduced
a new acquisition function \eqref{eq:Xsearch} grounded in the connection between the true optimum of the process $f$ and the expected number of crossings through its current estimate (cf. \eqref{eq:sup_prob}). However, such acquisition does not explicitly have into account any budget of failures $B$ or evaluations $T$. In the following, we propose an algorithm that makes use of $B$ and $T$ to balance the decision making between (i) safely exploring encountered safe areas, and (ii) searching outside the safe areas at the risk of failing, when safe areas contain no further information.

\subsection{Problem formulation}
To the unconstrained problem \eqref{eq:prob_uncons}, we add $G$ black-box constraints, $g_j : \dom \rightarrow \R$, $j=\{1,\ldots,G\}$, also
corrupted by noise and expensive to evaluate. Moreover, we assume a non-safety critical scenario, where violating the constraints is allowed, but it is strictly forbidden to do so more than $\budget$ times. Analogously, we allow only for a maximum number of $\evals \geq \budget$ evaluations. The case $\evals < \budget$ is not considered herein, as
the budget of failures can simply be ignored.
 Under these conditions, we formulate the constrained optimization problem with limited budget of failures as
\begin{align}
& \xmincons = \argmin_{x\in\dom}\; f(x),\text{   s.t.  } g_1(x) \leq 0,\ldots,g_G(x) \leq 0 \nonumber \\
& \text{under failures} \sum_{t=1}^\evals \Gamma(x_t) \leq \budget, \label{eq:prob_for}
\end{align}
where $\xmincons$ is the location of the constrained minimum, and $\Gamma(x_t) = \indi{g_1(x_t) > 0 \lor \ldots \lor g_G(x_t) > 0}$ equals 1 if at least one of the constraints is violated at location $x_t$, and 0 otherwise. $\mathbb{I}$ is the indicator function, and $g(x_1),\ldots,g(x_T)$ are the collected evaluations of the constraints
at locations $x_1,\ldots,x_T$.
Since the constraints $g_j$ are unknown, and modeled as independent Gaussian processes $g_j \sim \GP{0}{k(x,\hat{x})}$, queries $f(x)$ and $g(x)$ are stochastic and \eqref{eq:prob_for} cannot be solved directly. Instead, we address the analogous probabilistic formulation from \cite{Gelbart2014}: 
\begin{align}
& \xmincons \simeq \argmin_{x\in\dom}\;\mu(x), \text{s.t.  } \prod_{j=1}^G \Prob{g_j(x) \leq 0} \geq \rho \nonumber \\
& \text{under failures} \sum_{t=1}^\evals \Gamma(x_t) \leq \budget, \label{eq:original}
\end{align}
where $\Prob{g_j(x) \leq 0} = \CDF{-\mu_j(x)/\sigma_j(x)}$, $\Phi$ is the cumulative density function of a standard normal distribution, and $\rho \in (0,1)$ is typically set close to one. The predictive mean $\mu_j$ and variance $\sigma_j^2$ conditioned on $\mathcal{D}^{g_j}_t$ of each $g_j$ are computed as in \cref{ssec:GPs}.
In the following, we provide a novel Bayesian optimization strategy to address \eqref{eq:original}.
 
\subsection{Safe exploration with dynamic control}
\label{ssec:safe}
In order to include the probability of constraint satisfaction in the decision making, we propose a similar approach to \cite{Gelbart2014} by explicitly adding a probabilistic constraint to the search of the next evaluation
\begin{equation}
\begin{split}
\xnext = & \argmax_{\bm{x}\in\dom}\; \alpha_\text{X}(x)\\
& \text{s.t.  } \prod_{i=1}^K\Prob{g_j(x) \leq 0} \geq \rho_t,
\label{eq:optiprob}
\end{split}
\end{equation}
where the parameter $\rho_t \in (0,1)$ determines how much we are willing to tolerate constraint violation at each iteration $t$. This leads the search away from areas where the constraint is likely to be violated, as those areas get revealed during the search.

Contrary to \citep{Gelbart2014}, where $\rho_t$ is fixed a priori, we propose to choose it at each iteration, depending on the remaining budget of failures $\budgetRem = B - \sum_{j=1}^t\Gamma(x_t)$ and remaining iterations $\itersRem = T - t$. Intuitively, the more failures we have left (large $\budgetRem$), the more we are willing to tolerate constraint violation (large $\rho_t$). We achieve this by proposing an automatic control law to drive $\rho_t$, which we describe next.

Let us define a latent variable $z_t = \CDFinv{\rho_t}$, $z_t \in \R$ that follows a deterministic process $z_{t+1} = z_t + u_t$, using a dynamic feedback controller $u_t = u_t(\budgetRem,\itersRem)$.
Such controller
drives the process toward one of the two references: $\zsafe = \CDFinv{\dsafe}$ and $\zrisk = \CDFinv{\drisk}$, where typical values are $\dsafe = 0.99$ and $\drisk = 0.01$. We define a control law
\begin{equation}
u_t = (\zsafe - z_t)\tfrac{ \Gamma(x_t) }{\budgetRem} + (\zrisk - z_t)\tfrac{\budgetRem}{2\itersRem},
\label{eq:control}
\end{equation}
with $\budgetRem > 0$, $\itersRem > 0$, and $\budgetRem \leq \itersRem$. The first term drives the process toward $\zsafe$ when a failure occurs at iteration $t$, with intensity $1/\budgetRem$. In this way, the fewer failures are left in the budget, the more urgently the process chases $\zsafe$. The second term attempts to push $z_t$ down to $\zrisk$ with an intensity proportional to the ratio between the remaining failures and iterations.

When $\budgetRem = 0$, but $\itersRem > 0$, only a conservative safe exploration is allowed. To do so, we set $u_t = (\zsafe - z_t)$ for the remaining iterations until $t = T$. Additionally, if there are more failures left than remaining iterations, i.e., $\budgetRem > \itersRem$, the remaining budget of failures is not decisive for decision making, and thus, we set $u_t = (\zrisk - z_t)$.

The resulting control strategy weights risky versus conservative decision-making by considering the budget of evaluations and iterations left:
 When no failures occur for a few consecutive iterations, $\rho_t$ is slowly driven toward $\drisk$, and when a failure takes place, it lifts up $\rho_t$ toward $\dsafe$.

\subsection{Risky search of new safe areas}
\label{ssec:risky}

The probabilistic constraint in \eqref{eq:optiprob} puts a hard constraint on the decision making by not allowing evaluations in regions that are known to be unsafe. When $\rho_t$ is high, \eqref{eq:optiprob} will discard regions where no data has been collected and locally explore regions where safe evaluations are present. Such conservative decision making is desirable when $\budgetRem \ll \itersRem$ because it avoids unsafe evaluations. The smaller the $\rho_t$, the more risky evaluations we can afford, which makes the constraint information less important in the decision making. However, when $\rho_t$ is too low, the probabilistic constraint tends to be ignored, and the decisions are based on the information from the objective. Albeit this indeed counts as the wanted risky exploration strategy, completely ignoring the constraint information could result in repeated evaluations in unsafe areas. To avoid this, we follow the apporach from \citep{Gelbart2014}, where the aquisition function is aware of the constraint information, without this being a hard constraint. Therein, locations are chosen at
\begin{equation}
\xnext = \argmax_{\bm{x}\in\dom} \; \alpha_\text{X}(x)\prod_{j=1}^D\Prob{g_j(x) \leq 0}.
\label{eq:riskySearch}
\end{equation}
This approach ``jumps'' outside the current safe areas at the risk of failing, while 
the multiplying term discourages exploration in areas revealed to be unsafe.

Trading off risky versus safe exploration depends on the remaining budget $\budgetRem$, and is quantified by $\rho_t$, as detailed in \cref{ssec:safe}. We propose a user-defined decision boundary $\rho_\text{b}$, such that if $\rho_t \leq \rho_\text{b}$, the next location will be selected using \eqref{eq:riskySearch}, and \eqref{eq:optiprob} otherwise.

While \eqref{eq:optiprob} assumes that a safe area has already been found, this might not be the case at an early stage of the search. In such case, we collect observations using \eqref{eq:riskySearch} and only resort to the risk versus safety trade-off once a safe area has been found.

Pseudocode for the overall framework, named \emph{failures-aware excursion search} (\xsearchf), and an analysis of its computational complexity can be found in Appendix C. \xsearchf~returns
 the estimated location of the constrained minimum $\xmincons$ from \eqref{eq:original}, computed by setting $\rho = \dsafe$.

 \section{Empirical analysis and validation}
\label{sec:results}
We empricially validate \xsearch~and \xsearchf~by comparing their performance against state-of-the-art methods. We consider three different scenarios.
In the first one, we validate each method on common challenging benchmarks for global optimization. 
In the second and third scenarios we compare \xsearchf~against state-of-the-art methods in constrained optimization problems. In the second, we optimize the hyperparameters of a neural network to achieve maximum compression without degrading its performance. In the third, we learn the state feedback controller of a cart-pole system.
Both, \xsearch~and \xsearchf~are implemented in Python. The code, which includes scripts to reproduce the results presented herein, is documented and publicly available at \url{https://github.com/alonrot/excursionsearch}.

\subsection{Experimental setup}
To assess the performance of all methods we use \emph{simple regret} $r_T = f(x_\text{bo}) - \min_{x\in \dom} f(x)$, where $x_\text{bo} = \arg\min_{t \in [1,T]} y(x_t)$ is the point that yielded the best observation so far. In the constrained case, such point is given by $x_\text{bo} = \min_{t \in [1,T]} y(x_t) \text{ s.t. } y^g(x_t) \leq 0$. We quantify how often safe evaluations are collected using $\Omega = 100N_\text{safe}/T$, where $N_\text{safe}$ is the number of safe evaluations made at the end of each run.

In all cases, the domain is scaled to the unit hypercube. We set $\dsafe = 0.99$, $\drisk=0.01$, and $\rho_0 = 0.1$. The decision boundary was set at $\rho_\text{b} = 0.5$.
Both, the objective function and the constraint are modeled with a zero-mean GP, with a squared exponential kernel. The lengthscales and the signal variance are fit to the data after each iteration. Further implementation details, such as hyperprior choices and number of random restarts, are reported in Appendix D.

\subsection{Benchmarks for global optimization}
\label{ssec:benchmarksall}
We validate \xsearch~and \xsearchf~in two challenging benchmarks for global optimization: Hartman 6D, and Michalewicz 10D \citep{survey2013benchmarks}. 
We allow a budget of evaluations $T=100$ in all cases and repeat all experiments 50 times for each function using a different seed. As in \citep{wang2017mes,hernandez2016general}, we use the same initial evaluation (previously selected at random) across all repetitions.

\subsubsection{Excursion search (\xsearch)}
\label{sssec:benchmarksXs}
We assess the performance of \xsearch~by comparing against popular BO methods: Expected improvement (\ei) \citep{movckus1975bayesian},
Probability of improvement (\pim) \citep{kushner1964new}, Min-Value Entropy Search (\mes) \citep{wang2017mes}, and Gaussian process upper confidence bound (\ucb) \citep{Srinivas10gaussianprocess}. Our implementations are based on those used by \citep{wang2017mes}, available online\footnote{https://github.com/zi-w/Max-value-Entropy-Search}

\begin{figure}
\centering
\begin{subfigure}{0.49\columnwidth}
\includegraphics[width=\columnwidth]{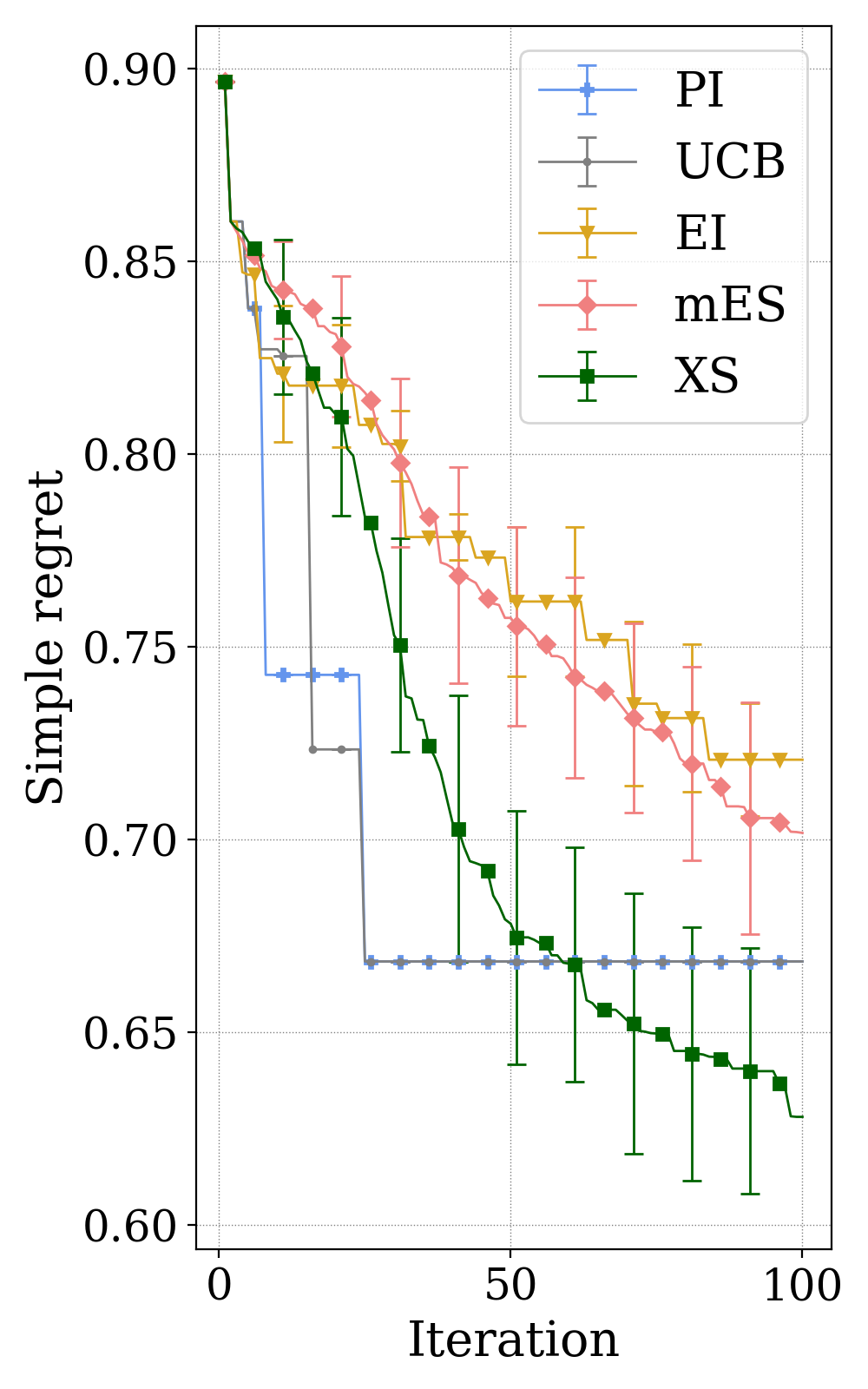}
\caption{Unconstrained}
\label{sfig:benchUnconsMicha10D}
\end{subfigure}
\begin{subfigure}{0.49\columnwidth}
\includegraphics[width=\columnwidth]{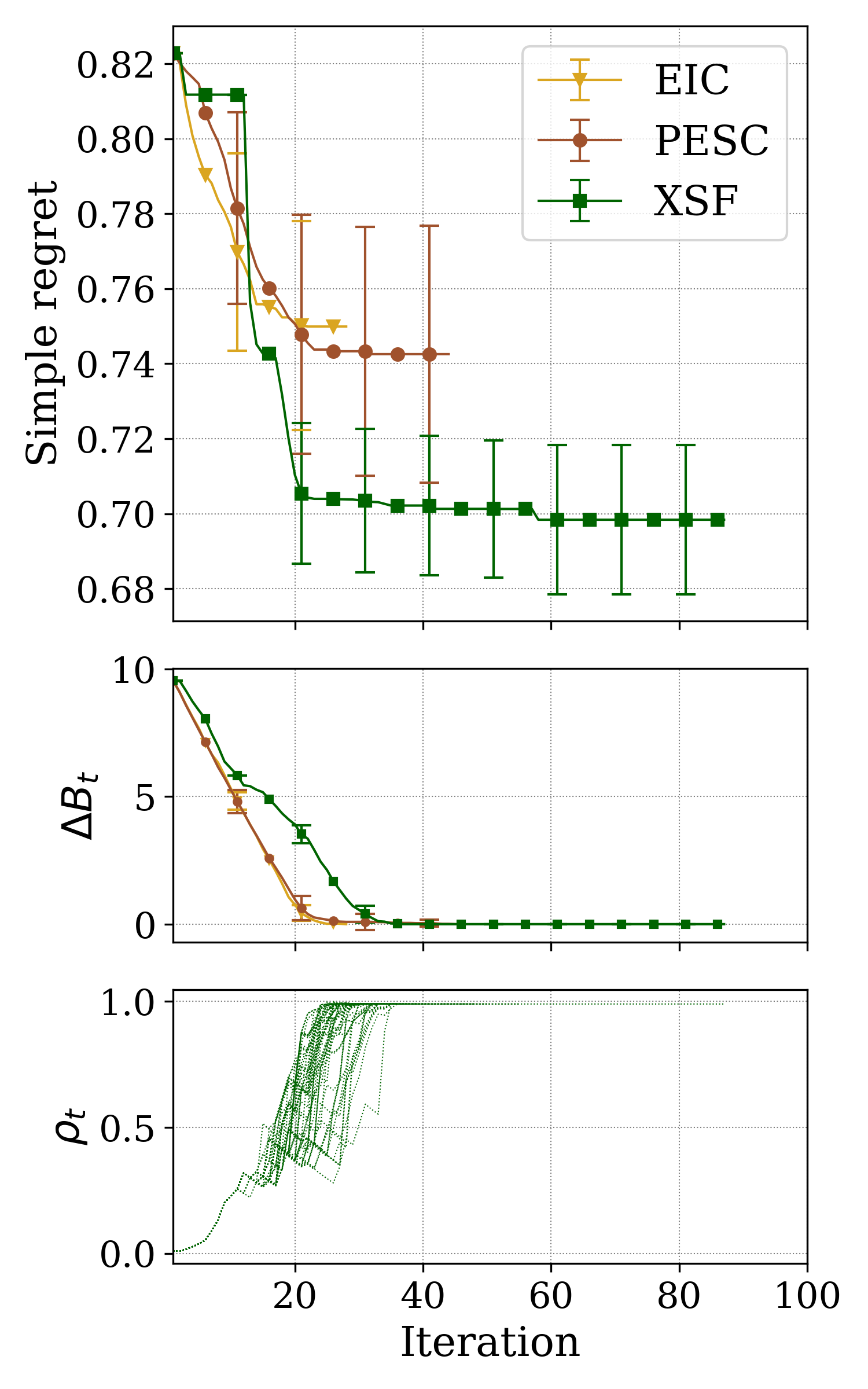}
\caption{Constrained}
\label{sfig:benchConsMicha10D}
\end{subfigure}
\caption{Performance assesment of \xsearch~and \xsearchf~on the Michalewicz 10-dimensional benchmark.}
\label{fig:benchConsAndUncons}
\end{figure}

\cref{sfig:benchUnconsMicha10D} shows the evolution of the simple regret over iterations in the Michalewicz 10D benchmark. \xsearch~reaches the lowest regret, and none of the methods is able to achieve a regret close to zero, which is not surprising given high dimensionality of the problem and the number of allowed evaluations.
\cref{tab:RegretBenchBothSafe} (top) shows statistics on the regret value for both benchmarks. While all methods report a generally high regret in Michalewicz 10D, \xsearch~clearly outperforms all the other methdos in Hartman 6D, as it finds a near-zero regret.

\tableRegretBenchBothSafe{b!}

\subsubsection{Failures-aware excursion search (\xsearchf)}
\label{sssec:benchmarksXsF}
To validate \xsearchf, we propose a constrained optimization problem under a limited budget of failures. For this, we simply impose a constraint to the aforementioned benchmarks $g(x) = \prod_{i=1}^D\sin(x_i) - 2^{-D}$.
Such function uniformly divides the volume in $2^D$ sub-hypercubes, and places $2^{D-1}$ convex disjoint unsafe areas in each one of the sub-hypercubes, so that they are never adjacent to each other. We allow $T=100$ and a considerably small budget of failures $B=10$ to all methods.
We compare \xsearchf~against expected improvement with constraints (\eic) \citep{Gelbart2014} and predictive entropy search with constraints (\pesc) \citep{hernandez2016general}. \eic~and \pesc~are terminated when their budget is depleted.
Although individual experiments rarely finish at the same iteration (i.e., some may deplete the budget of failures earlier than others),
we use in our results the last regret reported by each algorithm. For \eic, we use our own implementation, while for \pesc~we use the available open source implementation, included in Spearmint\footnote{https://github.com/HIPS/Spearmint/tree/PESC}.

In \cref{sfig:benchConsMicha10D}, we see that \xsearchf~reaches a higher number of total evaluations and consistently achieves lower regret than \eic~and \pesc. \cref{sfig:benchConsMicha10D} (middle) shows the evolution of the remaining budget of failures $\budgetRem$ over iterations (mean and standard deviation). As can be seen, 
\eic~and \pesc~deplete the budget faster than \xsearchf. Finally, \cref{sfig:benchConsMicha10D} (bottom) shows the evolution of the $\rho_t$ parameter used to switch betwen risky and safe strategies in \xsearchf, and also as a threshold for probabilistic constraint satisfaction (cf. \cref{ssec:safe}). We  differentiate two stages: During the initial iterations $\rho_t$ is low, and thus, risky exploration is preferred, which allows \xsearchf~to quickly discover better safe areas. At the last iterations, when the budget is depleted, \xsearchf~keeps exploring conservatively the discovered safe areas, with $\rho_t = \dsafe$.

\cref{tab:RegretBenchBothSafe} (bottom) shows the regret for both, the Michalewicz 10D and the Hartman 6D functions in the constrained case. While the regret comparison is similar to the 10D case, the 6D case shows that \xsearch~clearly outperforms the other methods. The quantity $\Omega$ confirms that \xsearchf~visits safe evaluations more often than the other methods.

Generally, hyperparameter learning influences the performance of the algorithms. In Appendix E, we show experiments with fixed hyperparameters and a correct GP model, where \xsearch~and \xsearchf~outperform the aforementioned methods.

\subsection{Compressing a deep neural network}

Applying modern deep neural networks (NNs) to large amounts of data typically results in large memory requirements to store the learned weights. Therefore, finding ways of reducing model size without degrading the NN performance has become an important goal in deep learning, for example, to meet storage requirements or to reduce energy consumption.
Bayesian compression has been recently proposed as a mean to reduce the NN size: Given an NN architecture, an approximate posterior distribution $q$ on the NN weights is obtained by maximizing the evidence lower bound (ELBO), which balances the
expected log-likelihood of samples from $q$ and the 
theoretical compression size, as given by the KL divergence between $q$ and a prior distribution $p$ \citep{havasi2018minimal}.
A penalization factor $\beta$ can be used to scale the KL divergence to control
the final size of the NN.
Finding the value of $\beta$
that achieves the lowest compression size without significantly degrading NN performance is a challenging and expensive tuning problem. To alleviate the effort of tuning hyperparameters, Bayesian optimization is commonly used. Herein, we propose to minimize the validation error of the NN while keeping its size below a threshold, using constrained Bayesian optimization under a limited budget of failures. While in this example failing to comply with the size requirements is not catastrophic, collecting many failures is undesirable.

\begin{figure}
\centering
\begin{subfigure}{0.49\columnwidth}
\includegraphics[width=\columnwidth]{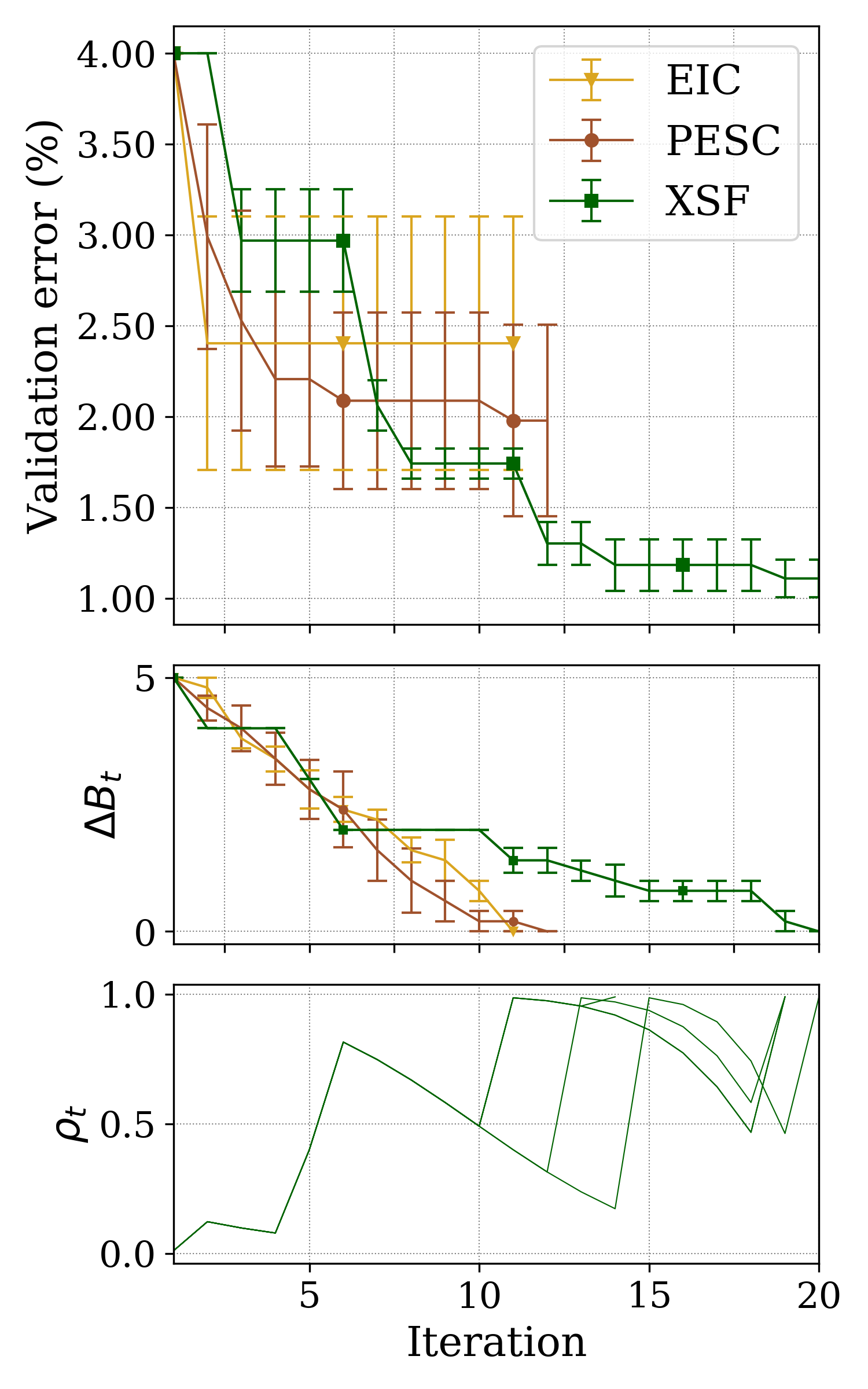}
\caption{NN compression}
\label{sfig:real_comp}
\end{subfigure}
\begin{subfigure}{0.49\columnwidth}
\includegraphics[width=\columnwidth]{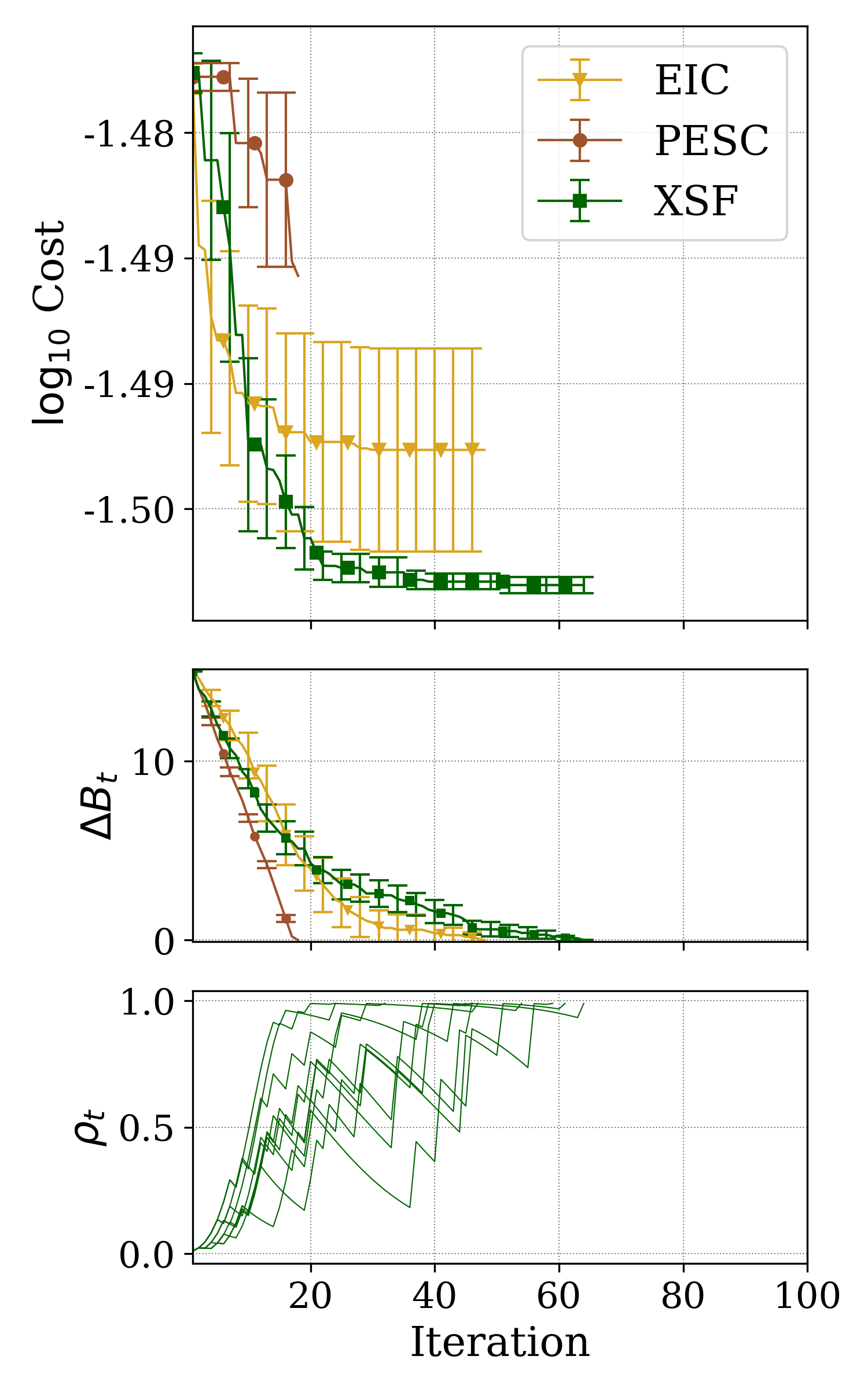}
\caption{Cart-pole problem}
\label{sfig:real_pend}
\end{subfigure}
\caption{Performance comparison of \xsearchf~against \eic~and \pesc}
\label{fig:real}
\end{figure}

We use a LeNet-5 on the MNIST dataset, and a required size below 15 kB. The parameters to tune are $\beta$, the learning rate $\chi$, and a scaling factor $\kappa$ on the the number of neurons of all layers. As a reference for our implementation, we used the open source implementation of MIRACLE\footnote{https://github.com/cambridge-mlg/miracle} \cite{havasi2018minimal}. We allow $T = 20$ and $B=5$ and repeat the experiments 5 times. 
We fix the training epochs to 20000 for each evaluation (about 25 min in wall-clock time). As shown in \cref{sfig:real_comp}, \xsearchf~achieves the lowest regret and standard deviation. 
The best safe observation is reported by \xsearchf, with validation error $\num{0.76} \%$ and theoretical NN size of $12.4$ kB (x553 compression). The learned parameters are $\beta=\num{6.56e-7}$, $\chi=\num{1.35e-3}$ and $\kappa=\num{10}$.

\subsection{Tuning a feedback controller}

Bayesian optimization has been used for learning robot controllers to alleviate manual tuning \citep{calandra2016bayesian,rai2018bayesian}. Herein, we propose to tune a 4D state feedback controller on a cart-pole system, where unstable controllers found during the search are undesirable, as human intervention is required to reset the platform, but not catastrophic. In this setting, allowing a limited budget of failures might increase chances of finding a better optimum. In practice, a constraint can be placed in the cart position to trigger an emergency stop when it grows large \cite{marco2016automatic}. Controllers that surpass such limit at any moment during the experiment are considered a failure. We use the simulated cart-pole system\footnote{https://gym.openai.com/envs/InvertedPendulum-v2/} from openAI gym \cite{openAI}, implemeted in the MuJoCo physics engine \cite{MuJoCo}. The tasks consists on, first stabilizing the pendulum starting from random initial conditions, and second, disturbing the cart position with a small step.
We consider a budget $B=15$ and $T=100$, and repeat all experiments 10 times. \cref{sfig:real_pend} shows that \xsearchf~finds a better controller than the other methods.

 \section{Conclusions}
\label{sec:conclusions}

In this paper, we have presented two novel algorithms for BO: Excursion search (\xsearch), which is based on the study of excursion sets in Gaussian processes, and failures-aware excursion search (\xsearchf), which trades off risky and safe exploration as a function of the remaining budget of failures through a dynamic feedback controller.
Our empirical validation shows that both algorithms outperform state-of-the-art methods. Specifically, in situations in which failing is permited, but undesirable, \xsearchf~makes better use of a given budget of failures.

\newpage
\clearpage
\bibliography{xsearch}
\bibliographystyle{icml2020}

\newpage
\clearpage
\appendix
\section{Additional details to Sec. 2.4}
\label{app:xsearch}
Herein, the derivation of (7) is complemented with two additional insights.
First, in \cref{sapp:analytical}, we show how the integral from (7) resolves into an analytical expression. Then, in \cref{sapp:virtual}, we reason about adding $\{x,u\}$ to the dataset $\sobs_t^f$ as a \emph{virtual} observation.

\subsection{Analytical expression for the integral in (7)}
\label{sapp:analytical}
The integral from (7) can be split in two parts
\begin{equation*}
\begin{split}
& \int_{-\infty}^{+\infty} |f^\prime| p(f^\prime|\tilde{\sobs})\text{d}f^\prime
=
-\int_{-\infty}^{0} f^\prime p(f^\prime|\tilde{\sobs})\text{d}f^\prime \\ 
& +\int_{0}^{+\infty} f^\prime p(f^\prime|\tilde{\sobs})\text{d}f^\prime,
\end{split}
\end{equation*}
where the placeholder $\tilde{\sobs} = \sobs_t^f \cup \{x,u\}$ is used for simplicity, and the dependency of $f^\prime$ on $x$ is implicit, and also omitted. Since $f^\prime\sim\N{f^\prime}{\mu^\prime(x),\nu^2(x)}$ is Gaussian distributed, each of the integrals above can be seen as the expected value of an unnormalized truncated normal distribution with support $[-\infty,0]$, and $[0,+\infty]$, respectively. These expectations are given by \citep{jawitz2004moments}
\begin{equation*}
\begin{split}
\int_{-\infty}^{0} f^\prime p(f^\prime|\tilde{\sobs})\text{d}f^\prime = \mu^\prime(x)Z_u(x) - \nu(x)\PDF{-\tfrac{\mu^\prime(x)}{\nu(x)}} \\
\int_{0}^{+\infty} f^\prime p(f^\prime|\tilde{\sobs})\text{d}f^\prime = \mu^\prime(x)Z_l(x) + \nu(x)\PDF{-\tfrac{\mu^\prime(x)}{\nu(x)}},
\end{split}
\end{equation*}
where $Z_l(x) = \CDF{\tfrac{\mu^\prime(x)}{\nu(x)}}$, $Z_u(x) = \CDF{\tfrac{-\mu^\prime(x)}{\nu(x)}}$, $\phi$ is the density of a standard normal distribution and $\Phi$ is its cumulative density function. We make use of the definition $\CDF{a} = \tfrac{1}{2}(1 + \erf{a/\sqrt{2}})$, where $\erf{\cdot}$ is the error function, to compute $\CDF{a} -\CDF{-a} = \erf{a/\sqrt{2}}$. Then, $Z_l(x)-Z_u(x) = \erf{\tfrac{\mu^\prime(x)}{\sqrt{2}\nu(x)}}$, and the integral can be solved analytically as
\begin{equation*}
\begin{split}
& \int_{-\infty}^{+\infty} |f^\prime| p(f^\prime|\tilde{\sobs})\text{d}f^\prime \\
& = \mu^\prime(x)(Z_l(x) - Z_u(x)) + 2\nu(x)\PDF{\tfrac{\mu^\prime(x)}{\nu(x)}} \\
& = \mu^\prime(x)\erf{\tfrac{\mu^\prime(x)}{\sqrt{2}\nu(x)}} + 2\nu(x)\PDF{\tfrac{\mu^\prime(x)}{\nu(x)}}.
\end{split}
\end{equation*}
Then, (7) follows.

\subsection{Virtual observation $\{x,u\}$}
\label{sapp:virtual}
The posterior of the process derivative $p(f^\prime|x,u,\sobs_t^f)$ is a Gaussian density and can be seen as conditioning $f^\prime(x)$ on an extended dataset that includes $\{x,u\}$ as a virtual observation. In the following, we briefly discuss this.

Since differentiation is a linear operation, the derivative of a GP remains a GP \citep[Sec. 9.4]{Rasmussen2006Gaussian}. Furthermore, the joint density between a process value $f(x)$, its derivative $f^\prime(x)$ and the dataset $\{X,y\}$ is Gaussian \citep{wu2017bayesian}
\begin{equation*}
\begin{split}
& p(y,f,f^\prime|x,X) = \\
& \Nbig{
\begin{bmatrix} y \\ f \\ f^\prime \end{bmatrix}
}{
\begin{bmatrix} 0 \\ 0 \\ 0 \end{bmatrix}
,
\begin{bmatrix} \tilde{K}(X,X) & K(X,x) & K^\prime(X,x) \\ 
								K(x,X) & K(x,x) & K^\prime(x,x) \\ 
								K^\prime(x,X) & K^\prime(x,x) & K^{\prime\prime}(x,x)
\end{bmatrix}
},
\end{split}
\end{equation*}
where $\tilde{K}(X,X) = K(X,X) + \sigma_\text{n}^2 I$, $K^\prime(X,x) = \partial K(X,x) / \partial x$, $K^{\prime\prime}(x,x) = \partial^2 K(x,x) / \partial x^2$, and the prior mean of the GP is assumed to be zero. Then, the conditional $p(f^\prime|f,x,\sobs_t^f) = \N{f^\prime}{\mu^\prime(x;f),\nu^2(x)}$ is also Gaussian, and can be obtained using Gaussian algebra \citep[A. 2]{Rasmussen2006Gaussian}. The mean $\mu^\prime(x;f)$ depends on the random variable $f$ as
\begin{equation}
\begin{split}
& \mu^\prime(x;f) = \\
& \begin{bmatrix} K^\prime(x,X) & K^\prime(x,x) \end{bmatrix}
\begin{bmatrix} \tilde{K}(X,X) & K(X,x) \\ 
								K(x,X) & K(x,x) \\ 
\end{bmatrix}^{-1}
\begin{bmatrix} y \\ f \end{bmatrix}.
\end{split}
\label{eq:virtual}
\end{equation}

The seeked Gaussian density $\N{f^\prime}{\mu^\prime(x;u),\nu^2(x)}$ is obtained by replacing the value $f$ in the expression for the mean \eqref{eq:virtual}. Thereby, $\{x,u\}$ appears in \eqref{eq:virtual} as an additional \emph{virtual} observation at location $x$ added to the existing dataset $\{X,y\}$, in shorthand notation: $p(f^\prime|u,x,\sobs_t^f) = p(f^\prime|\sobs_t^f \cup \{x,u\} )$.

 \section{\frechet~distribution}
\label{app:frechet}

In this section, we present a brief analysis on why assuming a \frechet~distribution is more error prone in practice than using the Gumbel distribution, when it comes to model the distribution over the global minimum $p(\fmin)$. This analysis complements Sec. 3.1 in the paper.

\begin{figure}[t!]
\centering
\includegraphics[width=\columnwidth]{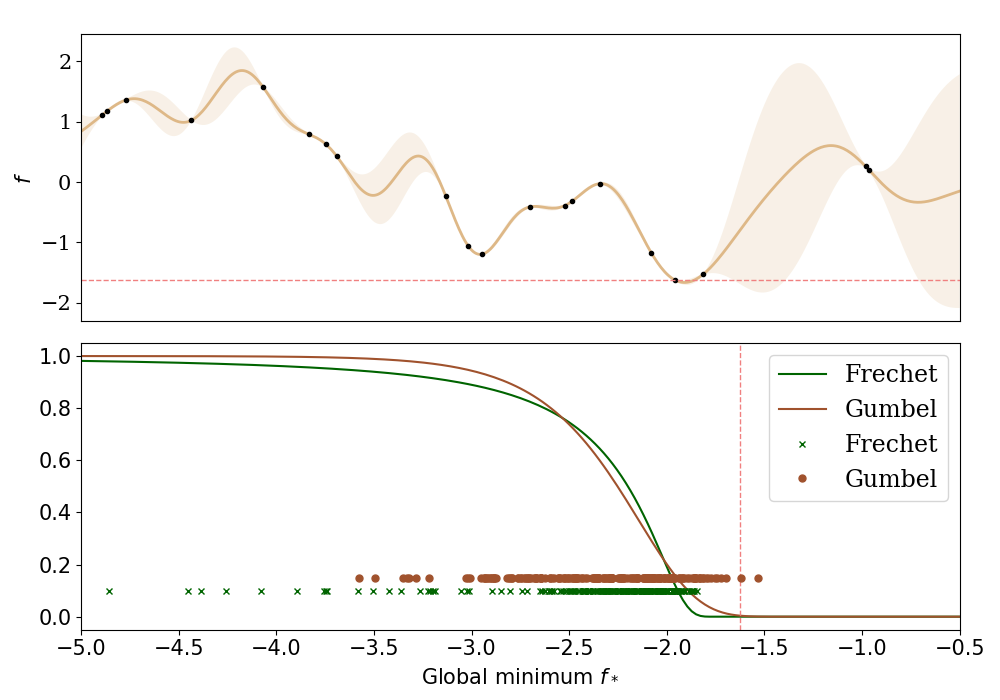}
\caption{(top) Gaussian process, and $\eta$ (red dashed line).
(bottom) Survival functions for both Gumbel, and \frechet~distributions. Samples from the Gumbel (crosses) and from the \frechet~(circles) distribution are shown. }
\label{fig:frechet}
\end{figure}

When modeling $p(\fmin)$ with the Gumbel distribution and sampling from it, some samples of the global minimum can lie above $\eta$, with non-zero probability, which is unrealistic. This can be explicitly avoided by using the \frechet~distribution which, contrary to Gumbel, has zero probability mass near $\eta$.
We illustrate this with an example, in which a GP with zero mean, unit variance, and squared exponential kernel is considered, conditioned on 20 observations sampled from the GP prior. We discretize the domain in 200 points and sample the resulting GP posterior at them. In \cref{fig:frechet}, we see that a portion of the Gumbel samples lie above $\eta$. To show consistency, we sample the posterior GP 100 times and average the number of times that Gumbel exceeds $\eta$, i.e., $1.60 \pm 1.22 \%$ of the cases, while the \frechet~distribution exceeds $\eta$ in $0 \%$ of the cases.

\section{Algorithm and complexity}
\label{app:algo}
Herein, we discuss pseudocode for \xsearchf~and its computational complexity.
\subsection{\xsearchf~algorithm}
Pseudocode for \xsearchf~is shown in \cref{alg:XSearch}. The decision boundary $\rho_\text{b}$ is used to switch between safe search (cf. (13)) and risky search (cf. (15)). The algorithm returns the location where the mean of the posterior GP is minimized without violating the probabilistic constraints. To abbreviate, we have used the placeholder $\varphi(x) = \prod_{i=1}^K\Prob{g_j(x) \leq 0}$.

We do not explicitly discuss \xsearch, as it simply comprises a standard Bayesian optimization loop, which involves (i) computing samples of the global minimum, and (ii) maximizing the acquisition function (10).

\subsection{Complexity}
At each iteration, the most expensive operations required to obtain (13) and (15) are: (a) obtaining samples from the global minimum $p(\fmin)$ and (b) maximizing the acquisition function using local optimization with random restarts.

As explained in \cite{wang2017mes}, obtaining $S$ samples from $p(\fmin)$ involves discretizing the input domain and performing a binary search, which has a total cost of $\mathcal{O}(S + N_d \log(1/\kappa))$, where $N_d$ is the size of the discretization grid, and $\kappa$ is the accuracy of the binary search. 

Each call to the acquisition function $\alpha_\text{X}$ (10), has a cost of $\mathcal{O}(SD)$ where $D$ is the dimensionality of the input space.
Then, assuming $R$ random restarts, and $M$ maximum number of function calls, the total cost of \xsearchf~in per iteration the worst case scenario is given by $\mathcal{O}(MRD(S+1) + N_d \log(1/\kappa) + (G+1)(N_\text{obs}+1)^3)$. The last term is the cost of inverting the Gram matrix, needed for GP predictions (cf. \eqref{eq:virtual}), after having collected $N_\text{obs}$ observations, and having $G$ constraints.
When setting $G=0$, we obtain the computational cost of \xsearch, as it also requires gathering samples from $p(\fmin)$ and local optimization with random restarts.

\renewcommand{\algorithmiccomment}[1]{\hfill #1}

\begin{algorithm}[tb]
	\caption{Failures-aware Excursion Search (\xsearchf)}
	\label{alg:XSearch}
	\begin{algorithmic}
	\STATE {\bfseries Input:} $T,B,\sobs_0^f, \sobs_0^g, \dsafe, \drisk, \rho_\text{b}, \rho_0$
		\FOR{$t=1$ {\bfseries to} $T$}
	\STATE $\rho_t \leftarrow$ {\sc UpdateDecisionBoundary}($\rho_{t-1}$) 	\STATE $\fmin \leftarrow$ {\sc SampleGlobalMinimum}($S$)
	\IF{$\rho_t > \rho_\text{b}$}
			\STATE $x_{t} \leftarrow \arg\max_{x\in \dom}\; \alpha_\text{X}(x;\fmin) \text{ s.t. } \varphi(x) \geq \rho_t $ \COMMENT{(13)}
	\ELSE
				\STATE $x_{t} \leftarrow \arg\max_{x\in \dom}\; \alpha_\text{X}(x;\fmin)\varphi(x)$ \COMMENT{(15)}
	\ENDIF
	\STATE {\sc EvaluateAndUpdateGPs}($x_t$)
	\ENDFOR
	\STATE $\xmincons \leftarrow \arg\min_{x\in \dom}\; \mu(x) \;\; \text{s.t. } \varphi(x) \geq \dsafe$
	\STATE {\bfseries Return:} $\xmincons$
			\vspace{3mm}
	\FUNCTION{{\sc UpdateDecisionBoundary}($\rho_t$)}
	\STATE $z_t \leftarrow \CDFinv{\rho_t}$
	\STATE $u_t \leftarrow u_t(\budgetRem,\itersRem)$ \COMMENT{Controller update (14)}
	\STATE $z_t \leftarrow z_t + u_t$ \COMMENT{Process update}
	\STATE {\bfseries Return:} $\CDF{z_t}$
	\ENDFUNCTION
			\vspace{3mm}
	\FUNCTION{{\sc SampleGlobalMinimum}($S$)}
	\STATE Estimate \frechet~distribution $\mathcal{F}_{s,q}$ following \citep[Appendix B]{wang2017mes}
	\FOR{$l=1$ {\bfseries to} $S$}
	\STATE $\fmin^l = \mathcal{F}_{s,q}^{-1}(\xi^l)$. $\xi^l \sim U(0,1)$
	\ENDFOR
	\STATE {\bfseries Return:} $\fmin^1,\ldots,\fmin^S$
	\ENDFUNCTION

	\vspace{3mm}
	\FUNCTION{{\sc EvaluateAndUpdateGPs}($x_t$)}
	\STATE $y=f(x_{t})$, $y_j=g_j(x_{t})\;j=\{1,\ldots,G\}$
	\STATE $\sobs_t^f \leftarrow \{y,x_{t}\}$, $\sobs_t^{g_j} \leftarrow \{y_j,x_{t}\}\;j=\{1,\ldots,G\}$
	\STATE Update hyperparameters of GP models
	\ENDFUNCTION
	\end{algorithmic}
\end{algorithm}

 \section{Implementation details}
\label{app:imple}

Both, \xsearch~and \xsearchf~are developed using {\sc BoTorch}\footnote{https://botorch.org/docs/introduction.html}, a Python library for Bayesian optimization that serves as a low-level API for building and optimizing new acquisition functions and fitting GP models. It makes use of {\sc scipy} Python optimizers\footnote{https://docs.scipy.org/doc/scipy/reference/tutorial/optimize.html} for estimating the GP hyperparameters and optimizing the acquisition function through local optimization with random restarts. In all cases we allow 10 random restarts and use {\sc L-BFGS-B} \citep{byrd1995limited} as local optimization algorithm. Currently, {\sc BoTorch} does not support optimization under non-linear constraints, which is needed to solve (13). To overcome this, we use the implementation of {\sc COBYLA} \citep{powell1994direct} from {\sc nlopt}\footnote{https://nlopt.readthedocs.io/en/latest/}.

In all experiments, the noise of the likelihood is fixed to $\sigma_\text{n} = 0.01$ for all GPs. The chosen hyperpriors on the lengthscales and the signal variance are reported in \cref{tab:hyper}, where $\mathcal{U}(a,b)$ refers to a uniform prior on the interval $[a,b]$,
$\mathcal{G}(a,b)$ refers to a Gamma prior with concentration $a$ and rate $b$, 
and $\mathcal{N}(a,b^2)$ refers to a normal distribution with mean $a$ and standard deviation $b$.

In Sec. 5.2., both, the Michalewicz and the Hartman functions are normalized to have zero mean and unit variance. The true minimum is known for both functions, which allows to compute the regret.

In Sec. 5.4, the goal is to find the state feedback gain $x \in \mathbb{R}^{4\times 1}$ for the cart-pole problem that minimizes a quadratic cost $f(x)$, which penalizes deviations of the pendulum states $s_k = [\varphi_k, \dot{\varphi}_k, l_k, \dot{l}_k]^\top$ from an equilibrium point $s^*$. 
The pole angle is $\varphi_k$, the pole angular velocity is $\dot{\varphi}_k$, the cart displacement is $l_k$, and the cart velocity is $\dot{l}_k$. 
The input to the system is the cart acceleration $a_k$, which is given by $a_k = x^\top(s_k - s^*) + 0.01\sum_1^{N_\text{simu}} (l_k - l^*)$, where an integrator, with gain $0.01$, is added to eliminate the steady-state the error.
For each parametrization $x$, the constraint value is computed as the maximum displacement of the cart over a simulation of $N_\text{simu}=800$ steps, i.e., $g(x) = \max(l_k),\; k = \{1,\ldots,N_\text{simu}\}$.
Constraint violation is quantified as $g(x) > l_\text{max}$, where $l_\text{max}$ is the physical limit of the rail in which the cart moves.
To allow the system to dissipate energy, the damping value of the simulated cart-pole in MuJoCo was increased from 1.0 to 1.5.\\\\

\tableHyper{h}

 \section{Additional results}
\label{app:results}
In this section, we present complementary results to Sec. 5.

To decouple the influence of the hyperparamater learning from the performance of the acquisition function itself, we fix the GP hyperparameters and sample the true objective $f$ and the true constraint $g$ from the corresponding GP priors. To obtain such samples we follow the same approach as in \citep{hernandez2016general}: First, the input domain is discretized to an irregular grid of 8000 points. Second, function evaluations are randomly sampled from the corresponding GP prior at such locations. Finally, the GP is conditioned on those evaluations and the resulting posterior mean is used as true objective. The lengthscales where fixed to $0.1$ and the signal variance to $1.0$.

The simple regret cannot be computed because the true minimum of the GP sample is unknown a priori. Instead, we report results assuming a very conservative lower bound on all the possible sampled functions, i.e., $\min_{x \in \dom}f(x) = -4.0$. 
We allow a maximum of $T=100$ iterations, and a budget of failures $B=15$ in the constrained case. The experiments were repeated 50 times for all algorithms. At each repetition, a new function is sampled from the GP priors.
 
In \cref{tab:RegretSynBothSafe},
we show a performance comparison of both, \xsearch~and \xsearchf~in optimizing a 3D input space. Without the influence of hyperparameter optimization, the proposed methods reach lower observations than state-of-the-art methods.

\tableRegretSynBothSafe{t!}

\end{document}